\documentclass[conference]{IEEEtran}
\IEEEoverridecommandlockouts
\usepackage{cite}
\usepackage{amsmath,amssymb,amsfonts}
\usepackage{algorithmic}
\usepackage{graphicx}
\usepackage{textcomp}
\usepackage{xcolor}
\def\BibTeX{{\rm B\kern-.05em{\sc i\kern-.025em b}\kern-.08em
    T\kern-.1667em\lower.7ex\hbox{E}\kern-.125emX}}
\begin{document}

\title{BAGS: An automatic homework grading system using the pictures taken by smart phones}

\author{
\IEEEauthorblockN{Xiaoshuo Li\IEEEauthorrefmark{1},  Tiezhu Yue\IEEEauthorrefmark{1}, Xuanping Huang\IEEEauthorrefmark{1}, Zhe Yang\IEEEauthorrefmark{2}, Gang Xu\IEEEauthorrefmark{3}}

\IEEEauthorblockA{\IEEEauthorrefmark{1}Beijing Boxfish Education Technology Co. Ltd, Beijing, China}
\IEEEauthorblockA{\IEEEauthorrefmark{2}Department of Physics, Tsinghua University, Beijing, China}
\IEEEauthorblockA{\IEEEauthorrefmark{3}School of Life Sciences, Tsinghua University, Beijing, China}
}

%\author{\IEEEauthorblockN{1\textsuperscript{st} Xiaoshuo Li}
%\IEEEauthorblockA{\textit{Boxfish} \\
%Beijing, China \\
%email address}
%\and
%\IEEEauthorblockN{2\textsuperscript{nd} Tiezhu Yue}
%\IEEEauthorblockA{\textit{Boxfish} \\
%Beijing, China \\
%email address}
%\and
%\IEEEauthorblockN{3\textsuperscript{rd} Xuanping Huang}
%\IEEEauthorblockA{\textit{Boxfish} \\
%Beijing, China \\
%email address}
%\and
%\IEEEauthorblockN{4\textsuperscript{th} Zhe Yang}
%\IEEEauthorblockA{\textit{Department of Physics} \\
%\textit{Tsinghua University} \\
%Beijing, China \\
%email address}
%\and
%\IEEEauthorblockN{5\textsuperscript{th} Gang Xu}
%\IEEEauthorblockA{\textit{School of Life Sciences} \\
%\textit{Tsinghua University} \\
%City, Country \\
%email address}
%}

\maketitle

\begin{abstract}
Homework grading is critical to evaluate teaching quality and effect. However, it is usually time-consuming to grade the homework manually. In automatic homework grading scenario, many optical mark reader (OMR)-based solutions which require specific equipments have been proposed. Although many of them can achieve relatively high accuracy, they are less convenient for users. In contrast, with the popularity of smart phones, the automatic grading system which depends on the image photographed by phones becomes more available. In practice, due to different photographing angles or uneven papers, images may be distorted. Moreover, most of images are photographed under complex backgrounds, making answer areas detection more difficult. To solve these problems, we propose BAGS, an automatic homework grading system which can effectively locate and recognize handwritten answers. In BAGS, all the answers would be written above the answer area underlines (AAU), and we use two segmentation networks based on DeepLabv3+ to locate the answer areas. Then, we use the characters recognition part to recognize students' answers. Finally, the grading part is designed for the comparison between the recognized answers and the standard ones. In our test, BAGS correctly locates and recognizes the handwritten answers in 91\% of total answer areas.
\end{abstract}


\section{Introduction}
Homework plays an important role in education. For teachers, by correcting students' homework, they can obtain feedback information on their teachings. For students, accurate correction of their homework in time is also necessary for their studies. However, it is usually time-consuming for teachers to grade the homework carefully and accurately. Therefore, developing an automatic homework grading method has attracted a lot of attention. Typically, the automatic homework grading method consists of three parts: the answer areas detection part, the answers recognition part, and the grading part to compare the recognized answers to the standard ones. Among these three parts, the answer areas detection part is the most important and difficult one.

\begin{figure}[t]
\begin{center}
\includegraphics[width=0.45\textwidth]{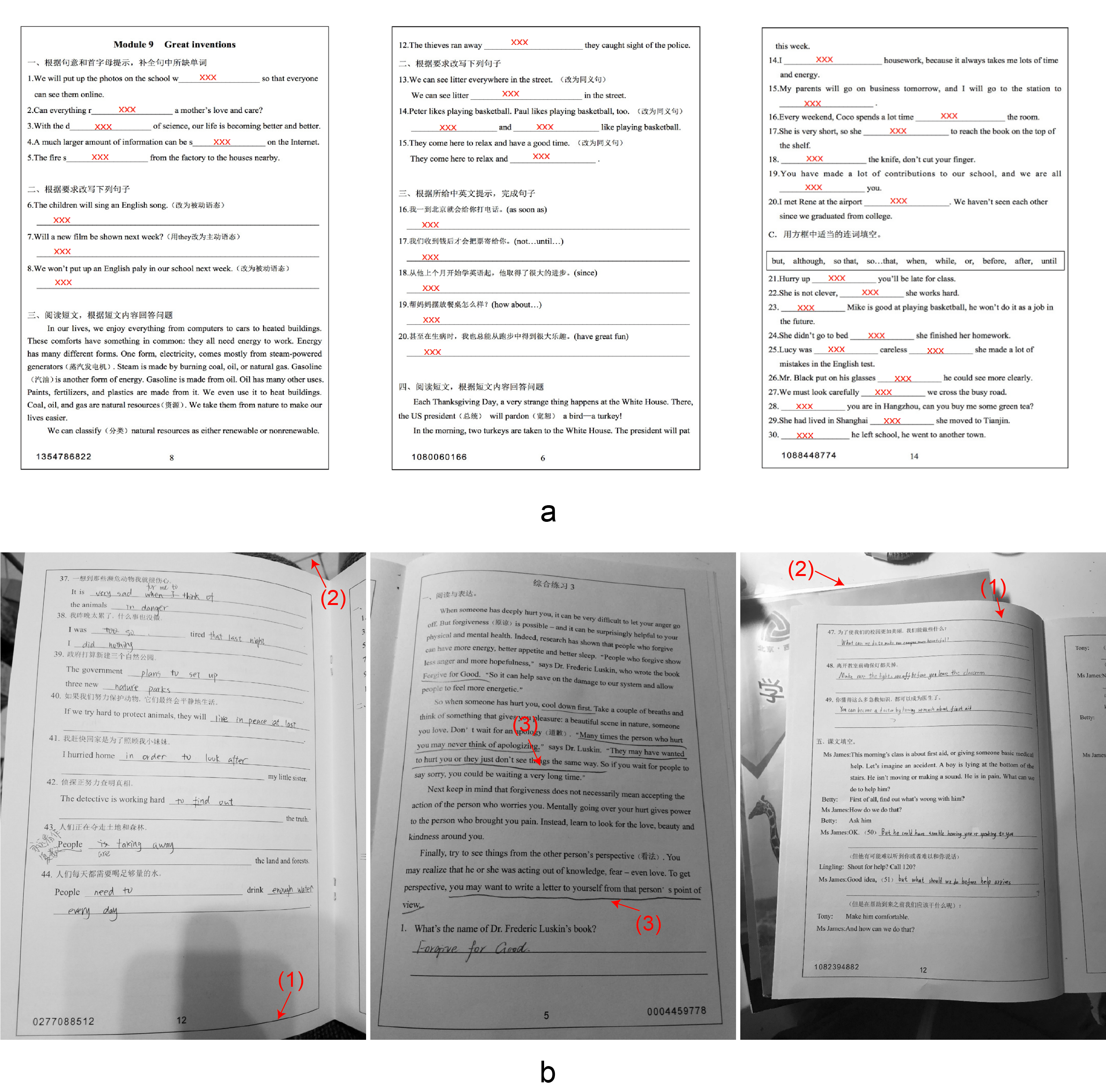}
\caption{The answer sheets in BAGS. a) shows some examples of the standard answer sheets. b) shows some images taken by smart phones and the answer areas detection difficulties on them: (1) is the distortion; (2) is the line-like texture in the background; (3) is the auxiliary line drawn by the students.}
\label{fig1}
\end{center}
\end{figure}

For traditional automatic grading methods, many of them are based on optical mark reader (OMR)~\cite{OMRsurvey}, therefore, they can locate the answer areas accurately. However, these methods depend on professional photographic equipments which are less convenient for the users~\cite{1981OMR,1999OMR}. In recent years, some works introduce the image processing techniques~\cite{2011OMR,2016OMR,Alomran2018OMR} to the OMR-based methods. Despite the great success of these methods, they are mainly focused on the multiple-choice questions. With the development of smart phones and their in-built camera technology, instead of relying on specific equipments for scanning, some automatic grading methods based on the pictures taken by smart phones have been proposed~\cite{2016Mobile,2018Boxes}. In this case, students, as well as their teachers, can get the grading results immediately without additional equipments. In practice, the answer areas detection is very challenging in these methods, and the challenges can be summarized as follows (see Fig.~\ref{fig1}(b)): (1) image distortion; (2) line-like texture in the background; (3) auxiliary lines drawn by students. Therefore, traditional lines detection algorithms such as Hough Transform~\cite{Hough} and line segment detection (LSD)~\cite{LSD} may not work well. In recent years, deep neural network has achieved many breakthroughs in semantic segmentation, making answer areas detection under this complex situation possible~\cite{FCN,Unet,deeplabv3,deeplabv3p}.

In this paper, we propose BAGS, an automatic homework grading system. For the convenience of the users and achieving real-time feedback, we allow users to take a picture of their homework using their phones and upload to BAGS. To solve the image distortion problem, all the answers would be written above the answer area underlines (AAU) and we use two semantic segmentation networks to locate these AAU. Compared to the traditional line detection methods, our segmentation method is more effective on our test set. After obtained the answer areas, we use the characters recognition part to recognize the handwritten answers and use the grading part to compare the recognized answers to the standard ones.

The contributions of this paper can be summarized as follows:
\begin{itemize}
  \item We propose an automatic homework grading system, BAGS, which can grade the homework photographed by smart phones. BAGS is convenient for the users and can get real-time feedback without additional equipments, which may benefit both students and teachers in the future teaching process.
  \item To segment thin objects (rectangular borderlines and answer area underlines (AAU) in this case), we modify the decoder in DeepLabv3+~\cite{deeplabv3p}, introducing the information when output stride is equal to 2 to recover more details. The performance of our model is better than that achieved by standard DeepLabv3~\cite{deeplabv3} and DeepLabv3+ on rectangular borderlines segmentation.
  \item In BAGS, we use semantic segmentation network to segment the AAU, which is better than the traditional line detection algorithms such as Hough Transform and line segment detection (LSD).
  \item To facilitate the researches in this area, most of our gathered datasets are available at https://github.com/boxfish-ai/BAGS.
\end{itemize}

\section{Related Works}
\subsection{Grade the Marked Answers}
The multiple-choice question is an important and widely used type of question, and it can be automatically graded by the optical mark reader (OMR)-based methods. In 1906, the first OMR system was developed by IBM depending on a set of light system, which can distinguish the different lights reflected by the marked areas and the unmarked areas~\cite{OMRsurvey}. In 1999, Chinnasarn and Rangsanseri~\cite{1999OMR} developed the first PC-based marking system depending on an ordinary optical scanner, which uses the image processing techniques for the answers recognition. In 2011, Nguyen et al.~\cite{2011OMR} successfully replaced the optical scanner by the camera for further convenience.

\subsection{Grade the Handwritten Answers}
Although the automatic grading system for the multiple-choice question has been widely studied, there are still many types of questions that require handwritten results. In 2016, I-Han Hsiao~\cite{2016Mobile} developed a novel mobile application to automatically grade the paper-based programming exams. However, in our case, more than one answer areas exist in a single image. In 2018, Amirali Darvishzadeh et al.~\cite{2018Boxes} proposed a method for the answer boxes detection which contains the handwritten answers and drawn by the users. However, as the number of questions increases, it becomes inconvenient for the users.

\subsection{Line Detection}
Traditionally, people use Hough Transform~\cite{Hough} and line segment detection (LSD)~\cite{LSD} to detect lines. However, due to the image distortion caused by different photographing angles or uneven papers, the line-like texture in the background, and the auxiliary lines from students (Fig.~\ref{fig1}(b)), these traditional methods may not work well. Recently, some deep convolutional neural network (CNN)-based semantic segmentation methods were used to detect the lines in images~\cite{2017Line,Xue2018Line,Huang2018Line}. Nan Xue et al.~\cite{Xue2018Line} transformed the problem of LSD as the region coloring problem because they share some similarities and the CNN-based segmentation methods are more accurate and efficient. Kun Huang et al.~\cite{Huang2018Line} presented the feasibility of junction and line segment detection in the man-made environments images using the respective neural networks.

\section{Methods}

\begin{figure*}[t]
\begin{center}
\includegraphics[width=0.95\textwidth]{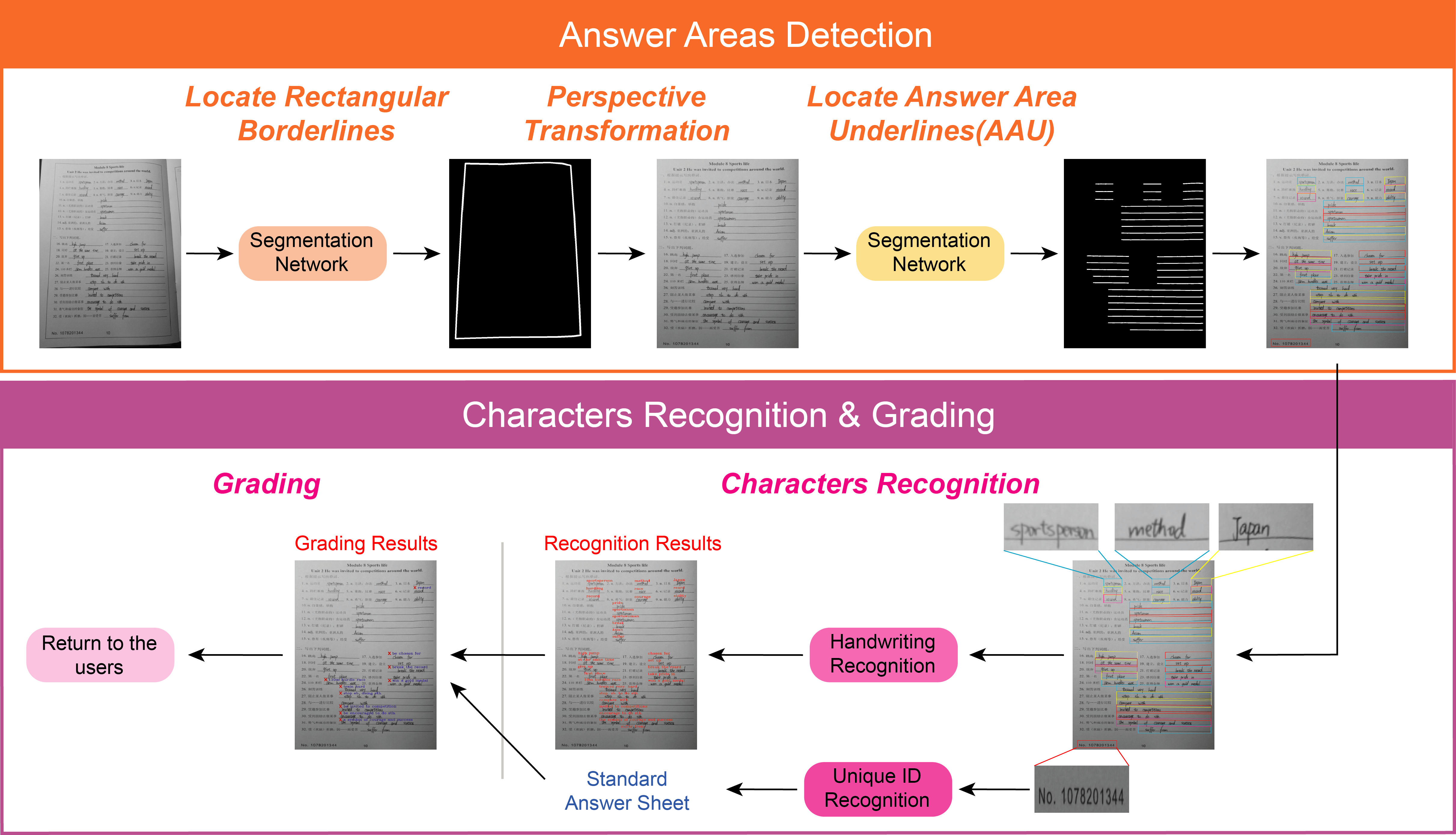}
\caption{The framework of BAGS. The colorful boxes are used to distinguish the different answer areas, the red font represents the recognition results, and the blue font represents the standard results. First, students take the picture of the answer sheet and upload to BAGS in grayscale for transfer efficiency. After automatic grading, the grading results are sent to the students immediately, achieving the real-time feedback. Meanwhile, the statistic report of the whole class can also be sent to the teachers for their teaching evaluations.}
\label{fig2}
\end{center}
\end{figure*}

As shown in Fig.~\ref{fig1}(a), for the convenience of the users, the questions and the answer areas are combined in the same \emph{answer sheet}, increasing the difficulty for the answer areas detection. Each answer sheet contains four \emph{rectangular borderlines} for image rectification, a \emph{unique ID} in the fixed position at left (or right) bottom for answer sheets distinction, and a corresponding \emph{standard answer sheet} which contains the standard results. Students need to write their answers above the \emph{answer area underlines} (AAU).

BAGS consists of three parts: the answer areas detection part, the characters recognition part, and the grading part (Fig.~\ref{fig2}). First, after receiving the pictures taken by the students, we use the answer areas detection part to locate the answer areas in the pictures. Then, based on their relative positions, we can obtain their corresponding standard results from the standard answer sheet. Second, we use the characters recognition part to recognize the handwritten results. Finally, we use the grading part to compare the recognized students' results with the corresponding standard results.

\subsection{Answer Areas Detection}

\begin{figure*}[t]
\begin{center}
\includegraphics[width=0.95\textwidth]{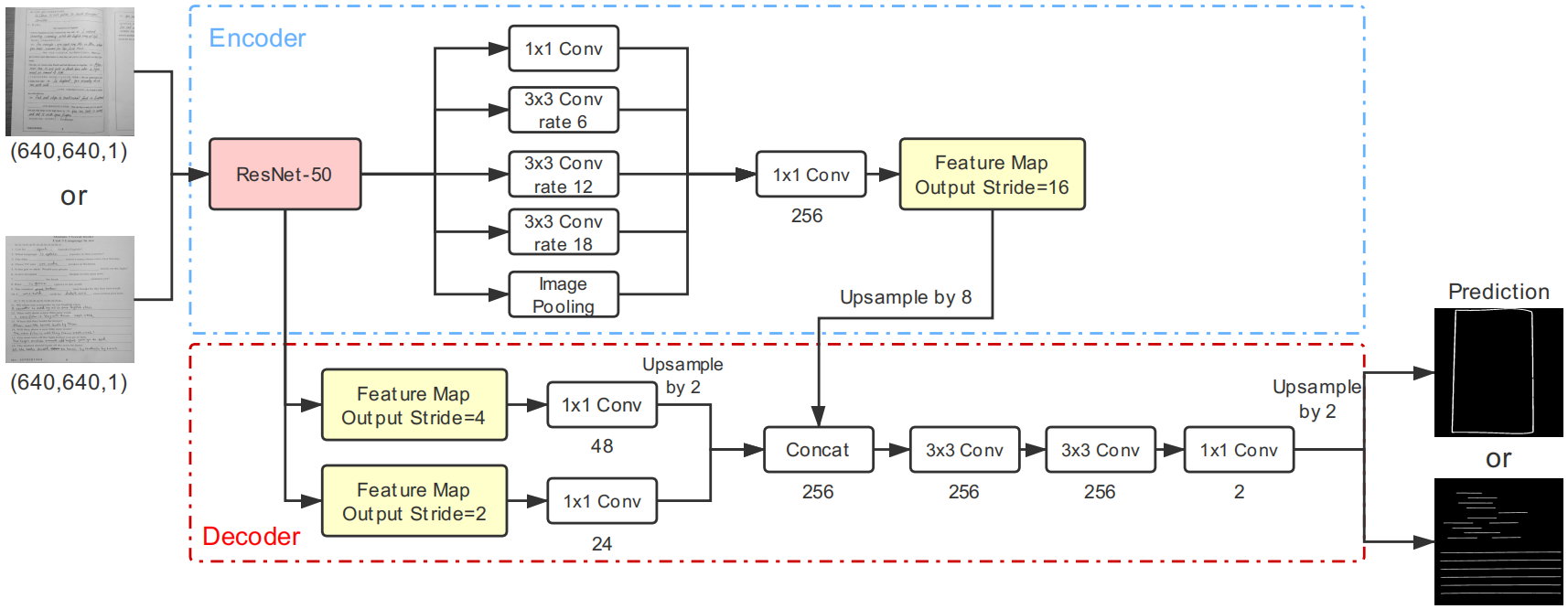}
\caption{The segmentation network in BAGS. The rate in the encoder stands for the atrous rate which is used to enlarge the receptive field of the filters. The numbers below convolutions are the numbers of filters. We train two segmentation networks with same structure to segment the rectangular borderlines and the answer area underlines (AAU), respectively. For the thin object segmentation, we add an extra route in the decoder of DeepLabv3+ to introduce the low-level features to the final prediction. Here, we use ResNet-50~\cite{resnet} as our backbone.}
\label{fig3}
\end{center}
\end{figure*}

\subsubsection{Basic Procedures}
The basic procedures of the answer areas detection part in BAGS can be described as follows (see Fig.~\ref{fig2}): Due to the issues we described in Fig.~\ref{fig1}(b), the traditional line detection methods may not work well. In this case, we use a semantic segmentation network to locate the rectangular borderlines at first. Then, we use Harris corner detection~\cite{harris} to locate four vertices of the rectangular borderlines and with which we rectify the image to the same size as the standard answer sheet. The distortion of the area that contains unique ID would be subtle because it is close to the left (or right) bottom vertex used in the perspective transformation. Therefore, we can directly put the area within the fixed position into the characters recognition part for the unique ID recognition, and search the corresponding standard answer sheet with it. For the answer areas detection, we choose to locate the AAU instead of the complete answer areas. We consider the areas where $m$ pixels above the AAU and $n$ pixels under the AAU as the complete answer areas, where $m$ and $n$ are set according to the size of standard answer sheet. Finally, we train another semantic segmentation network which has the same structure as the first one to locate the AAU. After obtained the AAU, we can get the topological information of them based on their relative positions, and get their corresponding standard results from the standard answer sheet.

\subsubsection{Semantic Segmentation Network}
To effectively locate the answer areas, we train two semantic segmentation networks with the same structure to segment the rectangular borderlines and the AAU, respectively. We segment them separately because the performance of the AAU segmentation is better when the image is rectified to the same size. Our segmentation network is mainly based on DeepLabv3+~\cite{deeplabv3p}, but with some modifications. Here, we denote \emph{output stride} as the ratio of input image spatial resolution to the final output resolution. For example, if the size of the input image is 640x640 pixels and the size of the feature map is 160x160 pixels, then the output stride is 4. In DeepLabv3+, the minimum output stride in its decoder is 4. Since lines are much thinner than the objects in traditional datasets~\cite{pascal}, we modify the decoder in DeepLabv3+, introducing the information when the output stride is equal to 2 to recover more details (Fig.~\ref{fig3}). Meanwhile, the input image is the grayscale image which only contains one channel for transfer efficiency, and the size of the input image is set to 640x640 pixels.

\subsection{Characters Recognition}
The characters recognition framework is performed using a CRNN network~\cite{crnn}. In BAGS, we train two networks with the same structure to recognize the printed unique ID and the handwritten results, respectively. Since the characters recognition is a relatively mature methodology, the details will not be further discussed.

\subsection{Grading}
After the first two parts, we obtained the students' results and the corresponding standard results. Therefore, the grading part can be easily implemented.

\section{Experiments}
\subsection{Data Preparation}
For the rectangular borderlines segmentation, we construct Dataset A, which consists of 9000 images for training and 1000 images for testing, and the images in Dataset A are the original grayscale pictures uploaded by the users. For the answer area underlines (AAU) segmentation, we construct Dataset B, which consists of 9000 images for training and 1000 images for testing, and the images in Dataset B are the rectified grayscale image. For the handwriting recognition, we construct Dataset C, which consists of 860000 images for training and 75000 images for testing. The images in Dataset C are the individual answer areas which include students' handwritten results, and they are labeled manually. To evaluate the answer areas detection and recognition performance of BAGS, we construct Dataset D. Dataset D consists of 383 images uploaded from 50 users which include 6068 answer areas, and the students' handwritten result in each answer area is manually labeled. All the images in these datasets are labeled and checked by three people to avoid mislabeling.

\subsection{Implementation}
We apply rotation, mirroring and scaling (between 1.0x and 1.05x) at random to augment the training data for two segmentation models. All the models in BAGS are implemented in TensorFlow \cite{tensorflow} and trained on one NVIDIA TITAN X. Both two segmentation models are trained using ADAM optimizer with a fixed learning rate of 0.001 for 20 epochs, and their batch sizes are set to 8. The handwriting recognition model is trained using SGD optimizer for 10 epochs. The learning rate is set to 0.1 at the beginning and then reduced by half after every 3 epochs. The batch size in handwriting recognition model is set to 128.

\subsection{Performance of Rectangular Borderlines Segmentation}
\begin{figure}[t]
\begin{center}
\includegraphics[width=0.45\textwidth]{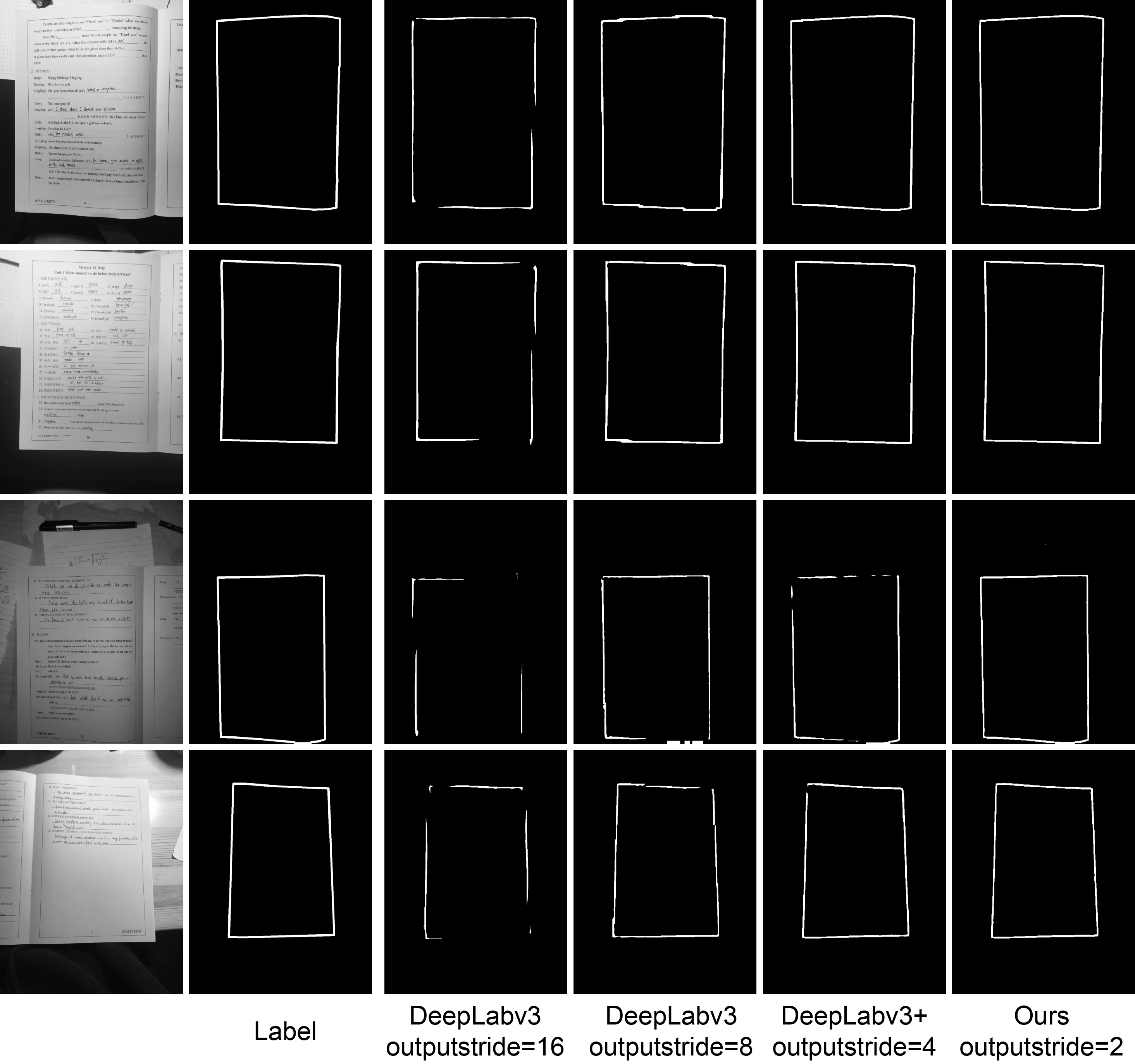}
\caption{Some examples of rectangular borderlines segmentation.}
\label{fig4}
\end{center}
\end{figure}

To segment the thin object (rectangular borderlines and answer area underlines (AAU) in this case), in BAGS, we modify the decoder in DeepLabv3+~\cite{deeplabv3p}, introducing the lower-level features information when the output stride is equal to 2 to recover more details. We compare our model with two DeepLabv3~\cite{deeplabv3} models which output stride are equal to 8 and 16, respectively and a standard DeepLabv3+ model which output stride is equal to 4. Here, all the models use ResNet-50~\cite{resnet} as their backbones. As Table~\ref{t1} and Fig.~\ref{fig4} show, in terms of pixel-level recall, precision, and accuracy, our model is better than others. Our results suggest that the information of low-level features is crucial for thin object segmentation. As the output stride goes smaller, the results go better (Table~\ref{t1}). It is noting that, although the uncomplete rectangular borderlines of another answer sheet also exist in the image, all the segmentation models choose to ignore them (Fig.~\ref{fig4}). In this respect, segmentation networks are better than traditional detection methods.

\begin{table}
\begin{center}
\caption{Pixel-level performance of rectangular borderlines segmentation using different models on Dataset A.}
\label{t1}
\begin{tabular}{|l|c|c|c|c|}
\hline
Model           &Output Stride  & Recall                & Precision             & Accuracy \\
\hline
DeepLabv3       & 16            & 46.5\%                & 77.2\%                & 98.0\% \\
DeepLabv3       & 8             & 80.5\%                & 88.3\%                & 99.1\% \\
DeepLabv3+      & 4             & 84.2\%                & 92.6\%                & 99.3\% \\
Ours            & 2             & {\bfseries 85.4\%}    & {\bfseries 94.4\%}  & {\bfseries 99.4\%} \\
\hline
\end{tabular}
\end{center}
\end{table}

\subsection{Performance of Answer Area Underlines (AAU) Segmentation}
The structure of the segmentation networks which are used to segment the rectangular borderlines and the AAU are the same. We separate these two processes because the performance of the AAU segmentation is better when the image is resized to a fixed size. We compare our AAU segmentation model with two traditional line detection methods: Hough Transform~\cite{Hough} and line segment detection (LSD)~\cite{LSD}. In Hough Transform, we first use the Sobel operator to detect the horizontal edges, the max gap of lines are set to 0, 5 and 10, respectively. In LSD, we retain the horizontal lines and the minimum length of lines are set to 20, 40 and 60, respectively.

\begin{table}
\begin{center}
\caption{Pixel-level performance of AAU segmentation using different methods on Dataset B.}
\label{t2}
\begin{tabular}{|l|c|c|c|c|}
\hline
Method           &Hyper-parameter  & Recall                & Precision             & Accuracy \\
\hline
Hough       & 0            & 30.4\%                & 69.0\%                & 97.0\% \\
Hough       & 5            & 35.5\%                & 66.9\%                & 97.1\% \\
Hough       & 10           & 40.9\%                & 53.3\%                & 96.6\% \\
\hline
LSD       & 20             & 66.5\%                & 57.2\%                & 97.1\% \\
LSD       & 40             & 64.8\%                & 63.7\%                & 97.5\% \\
LSD       & 60             & 63.1\%                & 64.4\%                & 97.4\% \\
\hline
Ours            & -             & {\bfseries 76.3\%}    & {\bfseries 70.5\%}  & {\bfseries 97.9\%} \\
\hline
\end{tabular}
\end{center}
\end{table}

\begin{figure}[t]
\begin{center}
\includegraphics[width=0.45\textwidth]{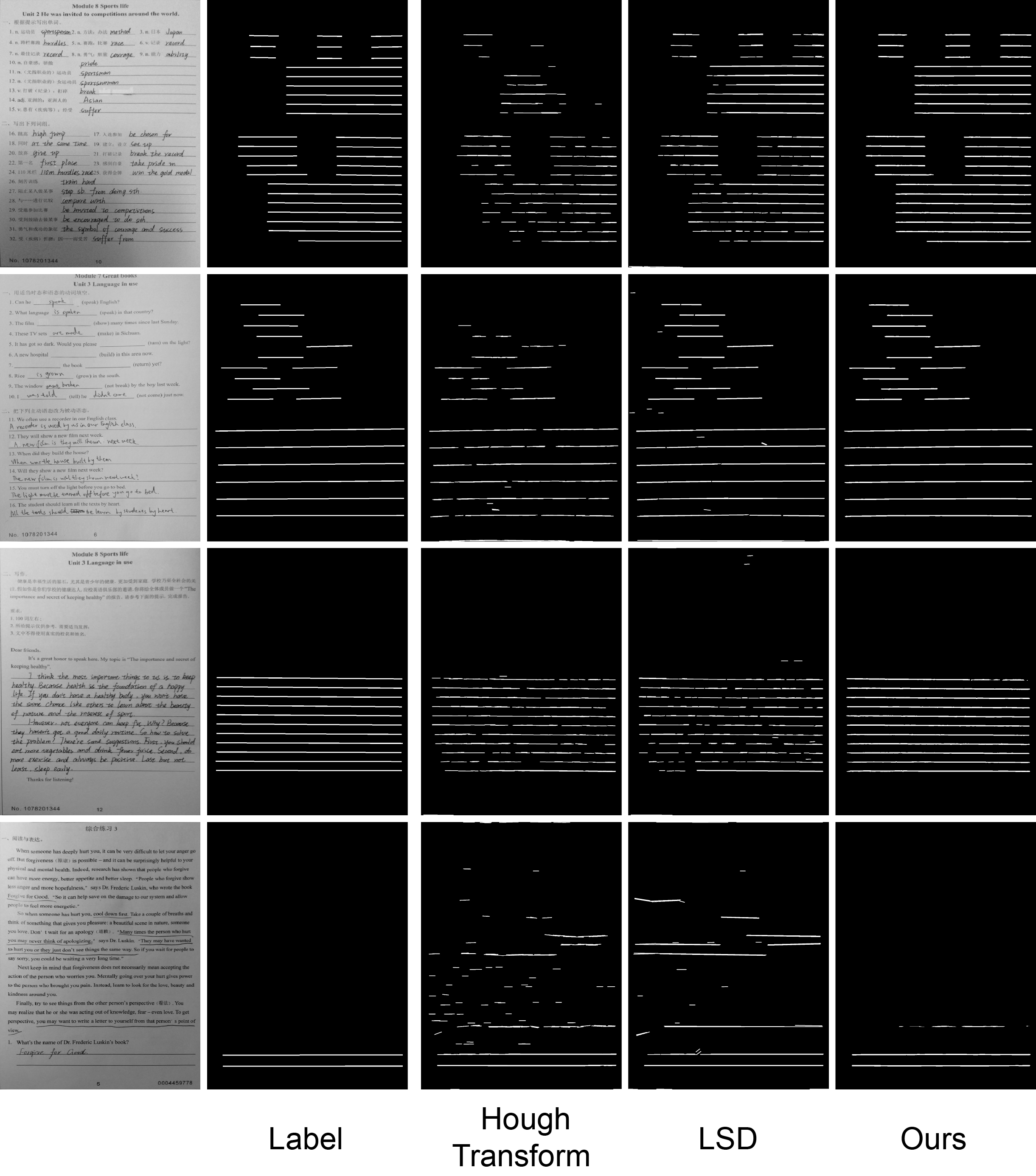}
\caption{Some examples of AAU segmentation. Here, the max gap of lines in Hough Transform is set to 5 and the minimum length of lines in LSD in set to 40.}
\label{fig7}
\end{center}
\end{figure}

As Table~\ref{t2} and Fig.~\ref{fig7} shown, our method is more effective than traditional line detection methods. The last example in Fig.~\ref{fig7} shows that, although there are many horizontal lines drawn by the users, our method is the least affected. In practice, such fewer misidentification lines can be easily corrected by the alignment of the answer areas in the standard answer sheet.

\subsection{Performance of BAGS}
We test our automatic grading system BAGS on Dataset D. Since the grading part can be easily implemented, we exclude the grading part and only focus on the students' results detection and recognition parts. In this test, if the unique ID is not located and recognized correctly, the accuracy of the whole answer areas in the corresponding image would be 0. Also, if the answer area is not located correctly, the accuracy of the corresponding answer area would be 0. On the contrary, if the answer area is correctly located, we calculate the accuracy between the recognized results and the label of corresponding answer area. Here, the accuracy is defined as follow:

\begin{small}
\begin{equation}
Accuracy=\begin{cases}
0                                                               & \text{if failed,}  \\
1-\frac{Levenshtein \ Distance}{Total \ Characters \ of \ Label} & \text{otherwise,}
\end{cases}
\end{equation}
\end{small}

where $Levenshtein \ Distance$ is a string metric for measuring the difference between two sequences~\cite{Levenshtein}.

\begin{figure}
\begin{center}
\includegraphics[width=0.45\textwidth]{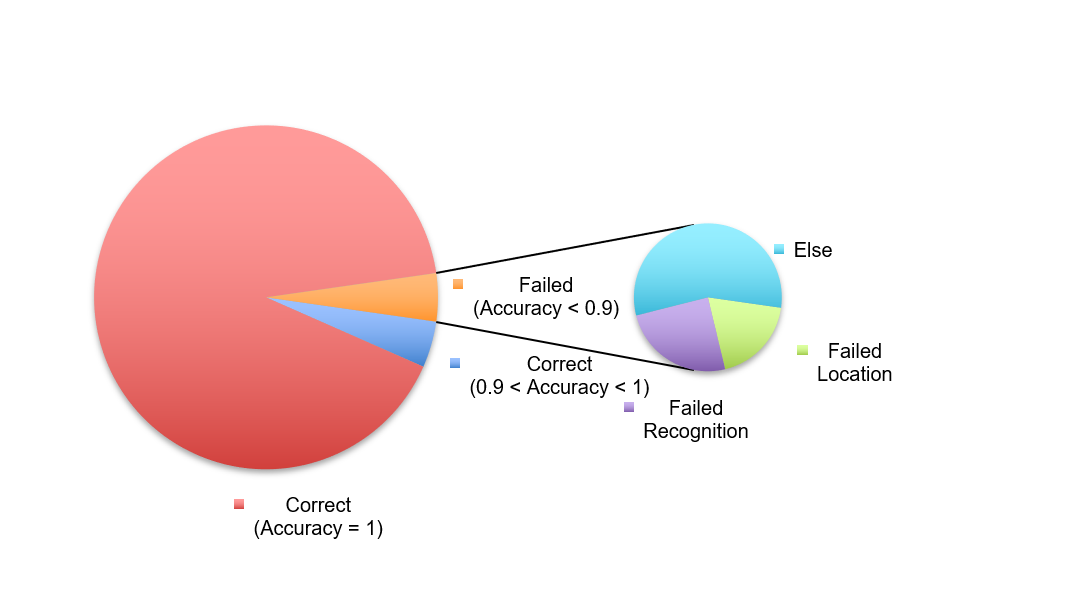}
\caption{Location and recognition performance of BAGS on Dataset D. \emph{Failed Location} represents the failures which caused by incorrect segmentation, \emph{Failed Recognition} represents the failures which caused by incorrect handwriting recognition, \emph{Else} represents the failures which caused by other reasons.}
\label{fig5}
\end{center}
\end{figure}

\begin{figure}
\begin{center}
\includegraphics[width=0.45\textwidth]{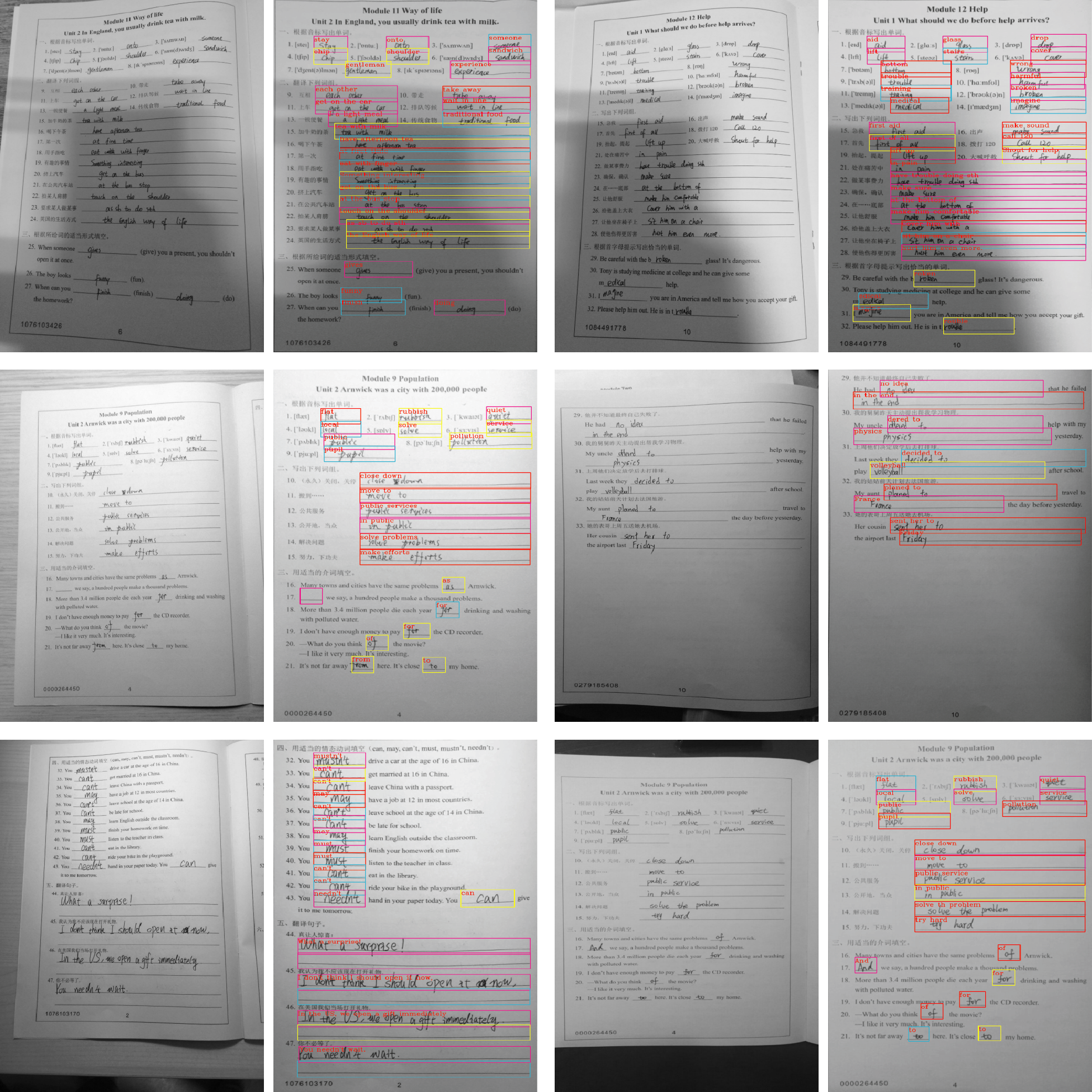}
\caption{Some segmentation and recognition results of BAGS. The colorful boxes are used to distinguish the different answer areas, and the red font represents the recognition results.}
\label{fig6}
\end{center}
\end{figure}

The Dataset D contains 383 images which include totally 6068 answer areas. The results in Fig.~\ref{fig5} show that for all the students' results in 6068 answer areas, above 91\% of them are completely correctly located and recognized ($accuracy = 1$), and less than 5\% of them are considered as failed ($accuracy < 0.9$). We further investigate the reason for the failures: in 274 failed images, 52 of which are caused by incorrect segmentation, 68 of which are caused by incorrect handwriting recognition, and 154 of which are caused by other reasons, such as ambiguous handwriting or low-quality images, etc. Fig.~\ref{fig6} shows some examples of our segmentation and recognition results.

\section{Conclusion}
In this paper, we introduce an automatic grading system, BAGS, which can grade the homework photographed by smart phones. We use two modified segmentation networks based on DeepLabv3+ to locate the answer areas under complex photographed backgrounds. We test BAGS on 383 images which include 6068 answer areas. BAGS correctly locates and recognizes above 91\% of them, and most of failures are caused by some other reasons, such as ambiguous handwriting or low-quality images. In BAGS, the procedures for rectangular borderlines segmentation and answer area underlines (AAU) segmentation are separated for higher accuracy because the later step requires images with the same scale, especially for the case where answer sheet constitutes only a small proportion of image. We will focus on this issue and simplify our pipeline in our future work. In summary, BAGS is an accurate and convenient automatic grading system which can achieve real-time feedback without using additional equipments, and this may benefit both students and teachers in the future teaching process.


\begin{thebibliography}{00}
%%%%%%%%%%%%%%%%%OMR %%%%%%%%%%%%%%
\bibitem{OMRsurvey}
Nirali P, Ghanshyam P. Various techniques for assessment of OMR sheets through ordinary 2D scanner: a survey[J]. Int J Eng Res Technol, 2015, 4(9): 803-807.
\bibitem{1981OMR}
Smith A M. Optical mark reading-making it easy for users[C]. Proceedings of the 9th annual ACM SIGUCCS conference on User services. ACM, 1981: 257-263.
\bibitem{1999OMR}
Chinnasarn K, Rangsanseri Y. Image-processing-oriented optical mark reader[C]. Applications of digital image processing XXII. International Society for Optics and Photonics, 1999, 3808: 702-709.
%%%%%%%%%%%%%%%%%OMR %%%%%%%%%%%%%%%

%%%%%%%%%%%% digital image process %%%%%%%%%%%
\bibitem{2011OMR}
Nguyen T D, Manh Q H, Minh P B, et al. Efficient and reliable camera based multiple-choice test grading system[C]. The 2011 International Conference on Advanced Technologies for Communications (ATC 2011). IEEE, 2011: 268-271.
\bibitem{2016OMR}
Bayar G. The use of hough transform to develop an intelligent grading system for the multiple choice exam papers[J]. Karaelmas Fen ve Mühendislik Dergisi, 2016, 6(1): 100-104.
\bibitem{Alomran2018OMR}
Alomran M, Chia D. Automated Scoring System for Multiple Choice Test with Quick Feedback[J]. International Journal of Information and Education Technology, 2018, 8(8).
%\bibitem{Smith2011OMR}
%Smith E H B, Goyal S, Scott R, et al. Evaluation of voting with form dropout techniques for ballot vote counting[C]. 2011 International Conference on Document Analysis and Recognition. IEEE, 2011: 473-477.
%%%%%%%%%%%% digital image process %%%%%%%%%%%

%%%%%%%%%%%%%% Apps in smart mobiles%%%%%%%%%%%%%%
\bibitem{2016Mobile}
Hsiao I H. Mobile grading paper-based programming Exams: Automatic semantic partial credit assignment approach[C]. European Conference on Technology Enhanced Learning. Springer, Cham, 2016: 110-123.
\bibitem{2018Boxes}
Darvishzadeh A, Entezari N, Stahovich T. Finding the Answer: Techniques for Locating Students' Answers in Handwritten Problem Solutions[C]. 2018 16th International Conference on Frontiers in Handwriting Recognition (ICFHR). IEEE, 2018: 587-592.
%%%%%%%%%%%%%% Apps in smart mobiles%%%%%%%%%%%%%%

%%%%%%%%%%%%%%%%% Line detection %%%%%%%%%%%
\bibitem{Hough}
Duda R O, Hart P E. Use of the Hough transformation to detect lines and curves in pictures[J]. Communications of The ACM, 1972, 15(1): 11-15.
\bibitem{LSD}
Von Gioi R G, Jakubowicz J, Morel J M, et al. LSD: a line segment detector[J]. Image Processing On Line, 2012, 2: 35-55.
%%%%%%%%%%%%%%%%% Line detection %%%%%%%%%%%

%%%%%%%%%%%% CNN%%%%%%%%%%%%%
% Resnet, FCN, Unet, DeepLab
\bibitem{FCN}
Long J, Shelhamer E, Darrell T. Fully convolutional networks for semantic segmentation[C]. Proceedings of the IEEE conference on computer vision and pattern recognition. 2015: 3431-3440.
\bibitem{Unet}
Ronneberger O, Fischer P, Brox T. U-net: Convolutional networks for biomedical image segmentation[C]. International Conference on Medical image computing and computer-assisted intervention. Springer, Cham, 2015: 234-241.
\bibitem{deeplabv3}
Chen L C, Papandreou G, Schroff F, et al. Rethinking atrous convolution for semantic image segmentation[J]. arXiv preprint arXiv:1706.05587, 2017.
\bibitem{deeplabv3p}
Chen L C, Zhu Y, Papandreou G, et al. Encoder-decoder with atrous separable convolution for semantic image segmentation[C]. Proceedings of the European Conference on Computer Vision (ECCV). 2018: 801-818.
%%%%%%%%%%%%%%%%CNN%%%%%%%%%%%%%

%%%%%%%%%%%%%%%%% Line detection %%%%%%%%%%%
\bibitem{2017Line}
Almazan E J, Tal R, Qian Y, et al. Mcmlsd: A dynamic programming approach to line segment detection[C]. Proceedings of the IEEE Conference on Computer Vision and Pattern Recognition. 2017: 2031-2039.
\bibitem{Xue2018Line}
Xue N, Bai S, Wang F, et al. Learning Attraction Field Representation for Robust Line Segment Detection[J]. arXiv preprint arXiv:1812.02122, 2018.
\bibitem{Huang2018Line}
Huang K, Wang Y, Zhou Z, et al. Learning to Parse Wireframes in Images of Man-Made Environments[C]. Proceedings of the IEEE Conference on Computer Vision and Pattern Recognition. 2018: 626-635.
%%%%%%%%%%%%%%%%% Line detection %%%%%%%%%%%

\bibitem{harris}
Harris C G, Stephens M. A combined corner and edge detector[C]. Alvey vision conference. 1988, 15(50): 10-5244.
%\bibitem{cityscapes}
%Cordts M, Omran M, Ramos S, et al. The cityscapes dataset for semantic urban scene understanding[C]. Proceedings of the IEEE conference on computer vision and pattern recognition. 2016: 3213-3223.
\bibitem{pascal}
Everingham M, Eslami S M A, Van Gool L, et al. The pascal visual object classes challenge: A retrospective[J]. International journal of computer vision, 2015, 111(1): 98-136.
\bibitem{resnet}
He K, Zhang X, Ren S, et al. Deep residual learning for image recognition[C]. Proceedings of the IEEE conference on computer vision and pattern recognition. 2016: 770-778.
\bibitem{crnn}
Shi B, Bai X, Yao C. An end-to-end trainable neural network for image-based sequence recognition and its application to scene text recognition[J]. IEEE transactions on pattern analysis and machine intelligence, 2017, 39(11): 2298-2304.
%\bibitem{alexnet} Krizhevsky A, Sutskever I, Hinton G E, et al. ImageNet Classification with Deep Convolutional Neural Networks[C]. neural information processing systems, 2012: 1097-1105.
%\bibitem{bilstm} Graves A, Fernandez S, Schmidhuber J, et al. Bidirectional LSTM networks for improved phoneme classification and recognition[J]. international conference on artificial neural networks, 2005: 799-804.
%\bibitem{ctc} Amodei D, Ananthanarayanan S, Anubhai R, et al. Deep speech 2: end-to-end speech recognition in English and mandarin[J]. international conference on machine learning, 2016: 173-182.
\bibitem{tensorflow}
Abadi M, Barham P, Chen J, et al. Tensorflow: A system for large-scale machine learning[C]. 12th {USENIX} Symposium on Operating Systems Design and Implementation ({OSDI} 16). 2016: 265-283.

\bibitem{Levenshtein}
Navarro G. A guided tour to approximate string matching[J]. ACM computing surveys (CSUR), 2001, 33(1): 31-88.

\end{thebibliography}
\end{document}